\documentclass[sigconf]{acmart}

\copyrightyear{2020}
\acmYear{2020}
\setcopyright{rightsretained}
\acmConference[AIES '20]{Proceedings of the 2020 AAAI/ACM Conference on AI,
Ethics, and Society}{February 7--8, 2020}{New York, NY, USA}
\acmBooktitle{Proceedings of the 2020 AAAI/ACM Conference on AI, Ethics, and
Society (AIES '20), February 7--8, 2020, New York, NY, USA}
\acmDOI{10.1145/3375627.3375832}
\acmISBN{978-1-4503-7110-0/20/02}

\usepackage
{color-edits}
\usepackage{multirow}
\usepackage{xspace}
\usepackage{tikz}
\usepackage{listings} 
\usepackage{enumitem}
\usepackage{array}
\usepackage{arydshln}

\usepackage{booktabs}
\usepackage{amsthm}

\newcommand{\inst}{\ensuremath{x}\xspace}
\newcommand{\Set}{\ensuremath{X}\xspace}
\newcommand{\type}{\ensuremath{a}\xspace}

\newcommand{\Types}{\ensuremath{A}\xspace}
\newcommand{\query}{\ensuremath{q}\xspace}

\newcommand{\Q}{\ensuremath{Q}\xspace}
\newcommand{\Z}{\ensuremath{Z}\xspace}
\newcommand{\inc}[4]{\ensuremath{\text{Inc}_{{#1}}(#2, #3,#4)}\xspace}
\newcommand{\inclusion}[4]{\ensuremath{\text{Inclusion}_{{#1}}(#2, #3,#4)}\xspace}
\newcommand{\diversity}[2]{\ensuremath{\text{Diversity}_{{#1}}(#2)}\xspace}
\newcommand{\presence}[2]{\ensuremath{\text{Presence}_{{#1}}(#2)}\xspace}
\newcommand{\dist}{\ensuremath{d}}

\newcommand{\rel}[2]{\text{rel}(#1,#2)}
\newcommand{\rep}[4]{\text{rep}_{#1}(#2, #3, #4)}
\newcommand{\I}{\mathbb{I}}

\AtBeginDocument{%
  \providecommand\BibTeX{{%
    \normalfont B\kern-0.5em{\scshape i\kern-0.25em b}\kern-0.8em\TeX}}}

\DeclareSymbolFont{extraup}{U}{zavm}{m}{n}
\DeclareMathSymbol{\varheart}{\mathalpha}{extraup}{86}
\DeclareMathSymbol{\vardiamond}{\mathalpha}{extraup}{87}

\addauthor{jm}{red} 
\addauthor{mm}{blue}
\addauthor{ed}{magenta}
\addauthor{nm}{purple}
\addauthor{ah}{cyan}
\addauthor{tg}{olive}
\addauthor{bh}{violet}
\addauthor{db}{orange}

\begin{document}
\fancyhead{}

\title{Diversity and Inclusion Metrics in Subset Selection}

\author{Margaret Mitchell}
\orcid{1234-5678-9012} 
\authornote{Corresponding author.}
\affiliation{%
  \institution{Google Research }}
\email{mmitchellai@google.com}

\author{Dylan Baker}
\affiliation{%
  \institution{Google Research}}
\email{dylanbaker@google.com}

\author{Nyalleng Moorosi}
\affiliation{%
   \institution{Google Research}}
 \email{nyalleng@google.com}

\author{Emily Denton}
\affiliation{%
\institution{Google Research }}
 \email{dentone@google.com}

\author{Ben Hutchinson}
\affiliation{%
 \institution{Google Research}}
\email{benhutch@google.com}

\author{Alex Hanna}
\affiliation{%
 \institution{Google Research }}
\email{alexhanna@google.com}

\author{Timnit Gebru}
\affiliation{%
  \institution{Google Research}}
\email{tgebru@google.com}

\author{Jamie Morgenstern}
\affiliation{%
   \institution{Google Research, University of Washington}}
\email{jamiemmt@google.com}

%

%
\begin{abstract}
The ethical concept of {\it fairness} has recently been applied in machine learning (ML) settings to describe a wide range of constraints and objectives. 
When considering the relevance of ethical concepts to subset selection problems, the concepts of {\it diversity} and {\it inclusion} are additionally applicable in order to create outputs that account for social power and access differentials.  
We introduce metrics based on these concepts, which can be applied together, separately, and in tandem with additional fairness constraints. Results from human subject experiments lend support to the proposed criteria. Social choice methods can additionally be leveraged to aggregate and choose preferable sets, and we detail how these may be applied.  
\end{abstract}

%
%


%


%

%
\maketitle

\section{Introduction}\label{sec:intro}
In human resource settings, it is said that {\it diversity} is being invited to the party; {\it inclusion} is being asked to dance \cite{HRThing}.  
Although difficult to define, such fundamentally human concepts are critical in algorithmic contexts that involve humans. Historical inequities have created over-representation of some characteristics and under-representation of others in the datasets and knowledge bases that power machine learning (ML) systems. System outputs can then amplify stereotypes, alienate users, and further entrench rigid social expectations. Approximating diversity and inclusion concepts within an algorithmic system can create outputs that are informed by the social context in which they occur.

In management and organization science, diversity focuses on
organizational demography; organizations that are diverse have
plentiful representation within race, sexual orientation,
gender, age, ability, and other identity aspects. 
Inclusion refers to a sense of belonging and ability to
function to one's fullest ability within organizations
\cite{mor1998tool,inclusion1,inclusion2,hope1999demographic}. In sociology, one strain of research assesses
the efficacy of diversity programs within firms, studying how well particular human resources interventions -- such as mentoring, anti-bias training, and shared organizational responsibility practices -- improve employee diversity \cite{kalev2006best,dobbin2016diversity}.  Another strain is skeptical of the concept of diversity
and the discursive work that it performs more broadly within firms and social life. Managers will often use the language of diversity without making corresponding changes to promote diverse and inclusive teams  \cite{bell2007diversity,embrick2011diversity,berrey2015enigma}.

An example of diversity is when people with different genders, races, and/or ability statuses
work together at a job. In this context, the people belong to
different identity groups. These identity groups are salient insofar
as they correspond to systems which afford them differential access to
power, as institutional racism, sexism, and ableism.  An example of inclusion is when wheelchair-accessible options
are available for wheelchair users in a building.  Here, the
wheelchair attribute is represented in the design of the building such
that wheelchair users are given similar movement options to those
without wheelchairs. Inclusion, in this case, refers to the ability of
individuals to feel a sense of both belonging and uniqueness for what
their perspective and abilities bring to a team \cite{inclusion1}.

Building on these concepts, we introduce metrics for {\it
  diversity} and {\it inclusion} based on quantifiable criteria that
may be applied in {\it subset selection} problems -- selecting a set
of instances from a larger pool. Subset selection is a common problem
in ML applications that return a set of results for a query, such as
in ranking and recommendation systems. While there are many burgeoning sets of mathematical formalisms for the
related concept of fairness, much of the work has focused on
formalizing anti-discrimination in the context of classification
systems. This has given rise to fairness criteria that call for parity
across various classification error metrics for pre-defined
groups \cite{barocas2016big}. Such constraints are generally referred to
as ``group" fairness, as they request that the treatment of each group
is similar in some measure. 
In contrast to group fairness, notions of individual fairness \cite{dwork2012fairness} ask that individuals similar for a task be treated similarly throughout that task.

Some notions of fairness proposed in the ranking and subset selection
literature include considerations that are closely related to the idea of 
diversity discussed here \cite{fair_diverse,assessing,fairness1,fairness2,fairness3}. However,
this literature has often conflated fairness and diversity as they are
referred to in other fields such as biology~\cite{bio1,bio2,bio3} and
ecology~\cite{ecology_commentary,ecology2,ecology3}. Geometric or
distance-based measures of diversity have also been explored within
the sciences, measuring the diversity of a dataset by the dataset's
volume~\cite{fair_diverse,7,13,14,11,23,24}, variance as in
PCA~\cite{samadi2018price}, or other measures of spread. The notion of {\it heterogeneity} more closely
matches such proposals, as they do not explicitly refer to features with
societal import and context.

Our work intentionally differentiates the concept of {\it diversity}
from {\it variety} or {\it heterogeneity} that may hold of a set,
where {\it diversity} focuses on individual attributes of social
concern (see the background section), and heterogeneity is agnostic to
specific social groups. As we discuss in this work, a diversity metric
can prioritize that as many identity characteristics as possible be
represented in a subset, subject to a target distribution. If the
target distribution is uniform (i.e., {\it equal representation}), this
is similar to demographic parity in fairness literature
\cite{dwork2012fairness}, where similar groups have similar
treatment. Although group-based fairness constraints may apply in this
setting, such constraints would be asking that all groups be
represented equally. The proposed diversity metrics allow for more control over
the specification of the distribution of
groups. 
Contrasted with the numerous definitions of diversity and
fairness, measurements of inclusion have
received relatively little consideration within computer science.  
We define a metric for inclusion, taking inspiration
from works in organization science and notions of individual fairness. To summarize our contributions:
\begin{enumerate}
\item We propose metrics for {\it diversity} and {\it inclusion}, relating these concepts to their corresponding social notions.
\item We focus on the general problem of selecting a set of instances from a larger set, formalizing how each set may be scored for diversity and inclusion. 
\item We demonstrate how methods from social choice theory can be used to aggregate and choose preferable sets.  
\end{enumerate}

Results from human subject experiments suggest that the proposed metrics are consistent with social notions of these concepts.

\section{Background and Notation}
\label{sec:background}

Subset selection is a fundamental task in many algorithmic systems, underpinning retrieval, ranking, and recommendation problems.
We formalize the family of diversity and inclusion metrics within this task. Fix a query \begin{math}\query\in \Q\end{math}, and a set of instances in the domain of relevance \begin{math}\Z_\query\end{math}.
\footnote{We intentionally conflate queries and query intents in this work, and assume that queries closely capture a user's intent.}

  Given a set of instances
  $\Set_{\query} \subset \Z_{\query}$ and instances $x_\query\in\Set_{\query}$, each instance $x_\query$
  may have multiple objects or items relevant to the query, e.g.,
  people or shoes. We denote these relevant objects by
  $x_{\query, i}$. All proposed metrics can act upon instances $x_q$ or
  sets $X_q$.

Let $\type$ refer to an \emph{attribute} of a person or item indexing a corresponding group type, such as {\it
  age:young}.  Here, the attribute \textit{young} indexes its
corresponding group type \textit{age}. $\type\in\Types$ defines the
set of attributes to measure for a given instance of set. With some
abuse of notation, we define $\type(\{p\})$ as a function that
indicates whether individual $p$ has attribute $\type$.  For example,
this might take the form of an indicator function.  We define
$\type(\{\inst\})$ as a function that indicates the relevance of
attribute $\type$ within $\inst$.  For example, this might take the
form of an indicator function for whether the instance contains an
item which refers to the attribute.  Similarly, we define
$\type(\Z_{\query})$ as a function of $\type$ within $\Z_{\query}$,
such as the proportion of instances $\inst_{\query}\in \Z_{\query}$
that contain $\type$.  This allows us to quantify the following
concepts for instances or sets:
\begin{itemize}
    \item[] {\bf Heterogeneity:} Variety within an instance or set of instances. $\Types$ may be any kind of characteristic, where greater heterogeneity corresponds to as many attributes $\type\in\Types$ in  $\Set_\query$ as possible.
    \item[] {\bf Diversity:} Variety in the representation of individuals in an instance or set of instances, with respect to sociopolitical power differentials (gender, race, etc.). Greater diversity means a closer match to a target distribution over socially relevant characteristics.
    \item[] {\bf Inclusion:} Representation of an individual user within an instance or a set of instances, where greater inclusion corresponds to better alignment between a user and the options relevant to them in an instance or set. 
\end{itemize}

 Throughout, we define $p$ as a set of attributes for an individual, but note that $p$ does not have to correspond to a specific person; it may simply be a set of attributes for a system to be inclusive towards. Critically, for the family of {\sc diversity} and {\sc inclusion}
metrics introduced below, $\Types$ is defined in light of human
attributes involved in social power differentials, such as gender,
race, color, or creed. Power differentials are significant insofar as
greater representation and presence of individuals with marginalized
identities can result in greater feelings of belonging and acceptance
and more successful teams. 
For example, if $\Types$ represents the Gender concept, an attribute
$\type\in\Types$ may be \{Gender:female, Gender:male, or
Gender:nonbinary\}. $\Types$ may also be a collection of attributes
from multiple different demographic subgroups, such as
\{Skin:Fitzpatrick Type 6, Gender:Female\}. Further details are
provided in the following section.

\section{Quantifying Diversity}\label{sec:diversity}

Recall the domain of relevance $\Z_\query$ for a query $\query$, and
the aim to quantify the diversity of a set
$\Set_\query\subset \Z_\query$.  The more diverse a set $\Set_\query$
is in a domain $\query$, the greater the presence of attributes
relevant to social structures of power and influence $\type\in\Types$ 
are represented in the set.

Given a set of attributes $\Types$ where each $\type\in\Types$ has
target lower and upper bounds on their presence in
$\type(\Set_{\query})\in[0,1]$ 
as a quantification of the presence of $\type$ within $\Set_{\query}$.
The measurement of $\type(\Set_\query)$ as well as the bounds
$l_\type$ and $u_\type$ are design parameters of our family of
diversity metrics. Selecting values for each induces a particular
metric in this family. The lower bound might be defined to implement
the $\frac{4}{5}$ rule, or require at least population-level frequency
of attribute $\type$ within $\Set$.  Many literatures have adopted
their own notions of diversity (see the introduction). Our formulation
bears some resemblance to that of \cite{CelisEtAl17}, who discuss
ranking objects subject to upper and lower bounds.  Our work departs
from theirs in that for different choices outlined below, these need
not be hard constraints on the presence of an attribute, and presence
need not implement simple count.
  
\paragraph{Presence Score} Recall that an instance $\inst_\query$
(e.g., a recommended movie in a set of movie recommendations) is
composed of one or more {\it items} (e.g., actors, objects, and
settings in the movie).  Each item reflects or indexes different
attributes. For example, the actors reflect attributes such as their
gender, age and race; objects similarly index such attributes, for
example, high heels may index the {\it woman}
attribute.
  We define the presence score of an attribute
$\type$ as a function quantifying how close the presence $\type(\inst_\query)$ is to the target and upper and lower bounds on the attribute's presence:
\[\presence{\type}{\inst_{\query}} = f(\type(\inst_{\query}),l_{\type},u_{\type})\]
with higher values meaning $\type$ is more present in
$\inst_{\query}$.

One natural quantification of the presence of $\type$ in
$\inst_\query$ is the proportion of items within $\inst_\query$
reflecting the attribute $\type$. Similarly, one of the simplest forms
that $f(\cdot)$ can take is as an indicator function that returns a
value of 1 when the the proportion of $\type$ in $x_\query$ is at
least $l_\type$. This approach is equivalent to:
$\presence{\type}{\inst_{\query}} = \I(1 \geq \type(\inst_{\query})
\geq l_{\type}|\inst_{\query})$. $f(\cdot)$ may also be instantiated
as a more complex function, for example, capturing the distance
between $\type(\inst_{\query})$ and $u_\type$.  There also may be
settings where the lower and upper bounds are not hard constraints:
some choices of $f$ can return nonzero values for
$\type(\inst_\query) \notin [l_\type, u_\type]$, such as when there is
an increasing penalty for going beyond the specified upper bound.

The presence formulation provides information about the contribution
of a single attribute to an instance. For each $\type$ the form of
$f(\cdot)$, as well as $\type(\cdot)$ $l_\type,u_\type$, must be
specified to define a metric. Different choices for these values give
rise to metrics with different meaning; what is appropriate for a
given task should be considered carefully by domain experts and a
broad set of individuals who use the technology relying on the set
selection.

Using target distributions for scoring sets and instances
provides for additional considerations beyond the parity often
afforded by fairness metrics, such as sets that are closer to
real-world distributions. This also potentially allows for more
fluid/nuanced treatment of group membership, where multiple
overlapping group memberships within one instance can be
accommodated. 

  \paragraph{Diversity Score}\label{sec:div_score} With the presence score defined, we can now
  define the diversity of an instance  $\inst_{\query}$ as an aggregate statistic of the attributes in the instance: 
  
  $\diversity{\Types}{\inst_\query} =
  g(\presence{\type}{\inst_{\query}})$, across $\type\in\Types$, where
  $g(\cdot)$ can return the minimum, maximum, or average presence
  value of the attributes.  These standard choices of cumulation
  functions are borrowed from social choice theory in economics, and
  similar economics-based metrics may be applied to combine presence
  scores of many attributes into the single diversity score, for
  example, using a function such as $\text{maximin}$ \cite{rawls1974some}
  reduces to the lowest-scoring attribute $\type$ for
  $\presence{\type}{\inst_\query}$ (see below section on Social Choice
  Theory).

The Diversity family of metrics can highlight or prioritize diversity
with respect to relevant social groups.  For example:
\begin{itemize}
    \item[]{\bf Racial Diversity:} many race groups $\type\in\Types$ present.
    \item[]{\bf Gender Diversity:} many gender groups $\type\in\Types$ present.
    \item[]{\bf Age Diversity:} many age groups $\type\in\Types$ present.
\end{itemize}

\begin{figure}
    \centering
    \includegraphics[height=1cm]{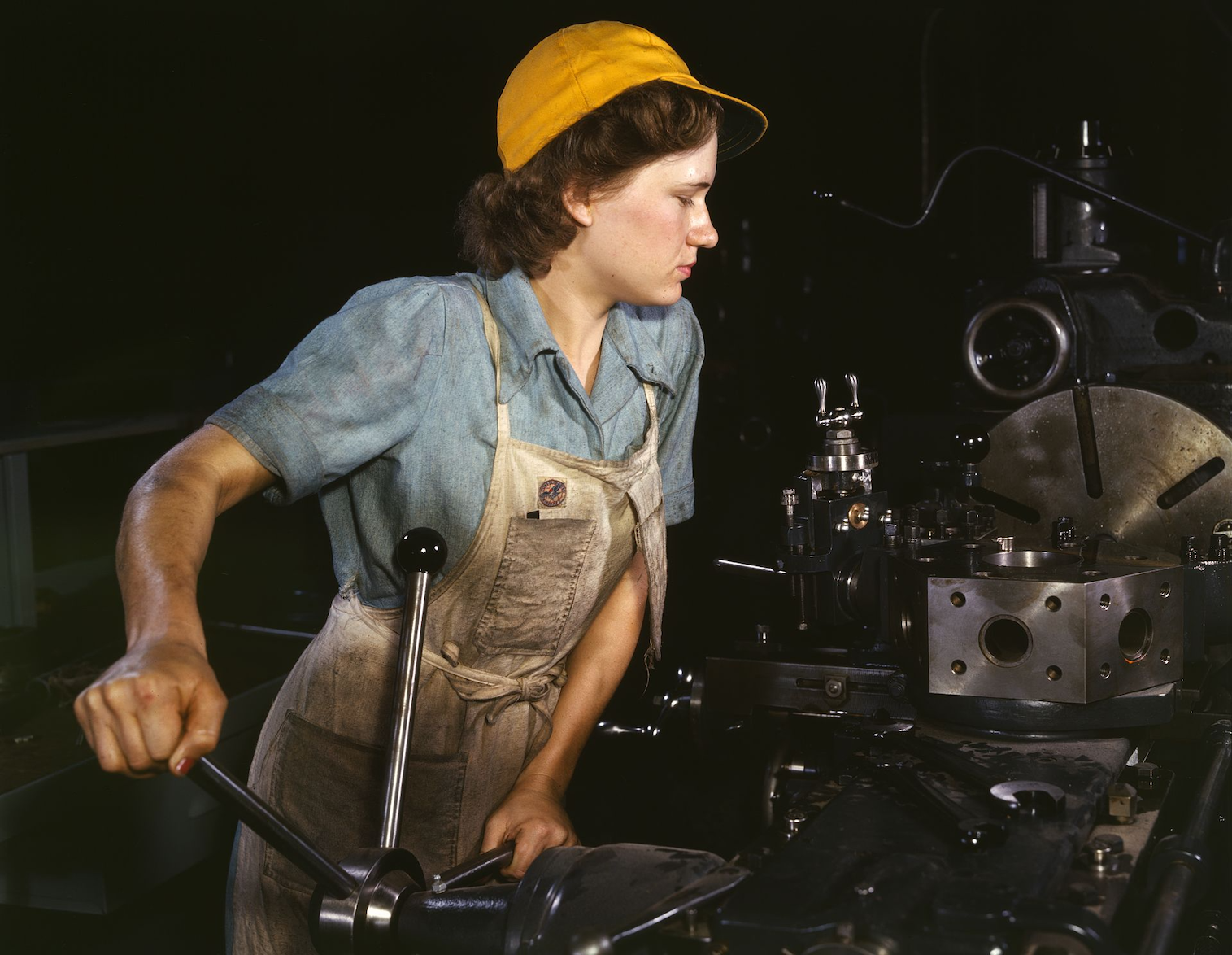}
    \includegraphics[trim=10 0 0 0,clip,height=1cm]{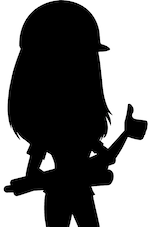}
    \includegraphics[height=1cm]{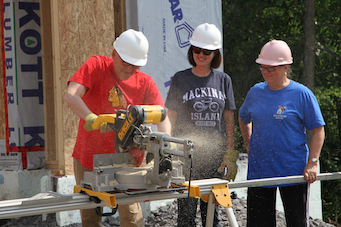}
        \includegraphics[height=1cm]{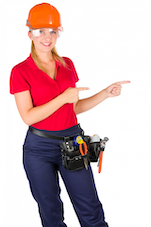}
        \includegraphics[height=1cm]{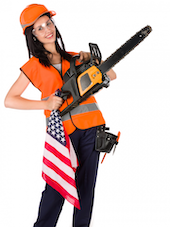}
    \includegraphics[height=1cm]{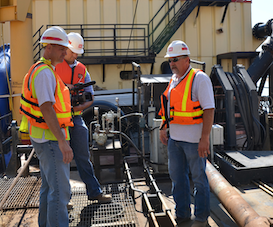}
    \includegraphics[height=1cm]{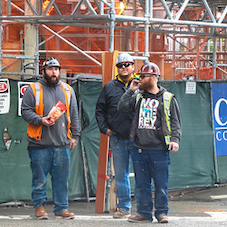}    
    \includegraphics[height=1cm]{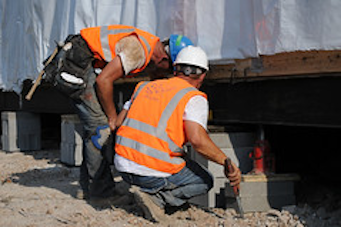}  
    \includegraphics[height=1cm]{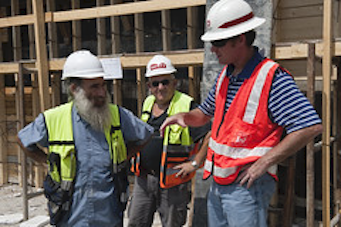}   
    \includegraphics[height=1cm]{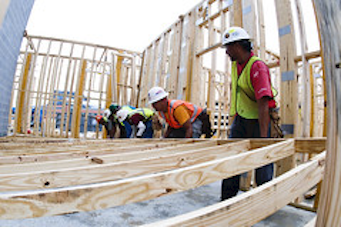}   
    \includegraphics[height=1cm]{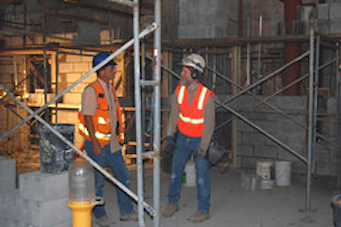}   
    \includegraphics[height=1cm]{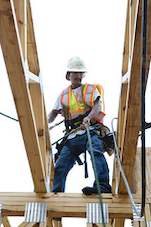}   
    \caption{Gender diversity, without inclusion for women, ``construction worker'' image domain: Although several genders and colors are represented (diversity of people), male-presenting individuals are shown in realistic, modern construction worker situations, while women and other genders are depicted as historic nostalgia, toys, clipart, or passive in the event.}
    \label{fig:div_ex1}
\end{figure}

\vspace{-1em}
\paragraph{Set Diversity} The formulation for an instance giving rise
to a diversity score naturally extends to a set of instances giving rise
to a diversity score.  An example set of images that are Gender Diverse are shown in Figure 1.  We define the {\it cumulative diversity score}
of a set $\Set_\query$ as a function of
$\diversity{\Types}{\inst_\query}$ across
$\inst_\query \in \Set_\query$. As before, this can be scored following the social choice theory 
functions further detailed below.

\section{Quantifying Inclusion}\label{sec:inclusion}

We now move towards proposing a family of metrics to measure inclusion
for subset selection.  Our proposed inclusion metric captures the
degree to which an individual $p$ is well represented by the returned
set. As an example, an individual looking for hair style inspiration
might query `best hairstyles 2019'. In the absence of additional
qualifiers, e.g., those that narrow the query by explicitly specifying
demographic information, an inclusive image set would be one where
the individual sees people with similar hair textures to theirs in
the selected set. We measure the inclusion of a person (or set of attributes) $p$ along attribute
$\type$ when selecting $\Set_\query$ from
$\Z_\query$. 
We begin by introducing {\it instance inclusion}, a measure of how
well an instance $\inst_\query$ represents $p$, and then extend to set
inclusion.

\paragraph{Instance Inclusion.}
As above, we assume an instance $\inst_\query$ (e.g., an image) is
composed of one or more \emph{items} (e.g., different components of
the image).  Each item has some relevance to a query $\query$ and may
be a better or a worse fit for an individual $p$ along some attribute
$\type$.  The inclusion of an instance $\inst_\query$ aggregates the
relevance and fit of all items in $\inst_\query$ and produces a
single measure of that instance's ability to reflect $p$ or to meet
$p$'s goals. 

Continuing with the example above, an instance can refer
to an image with several subjects, and each subject corresponds to an
item $i$. A person $p$ may find $i$ to be a good fit along the hair
type attribute if their hair type is similar to $p$'s. Then, the
instance's inclusion for $p$ along this attribute combines the fit of
all the subjects $i$ in the instance.

\paragraph{Relevance of an item.}
Formally, let $\rel{i}{\query}\in [0,1]$ measure the \emph{relevance} of
an item $i$ to query $\query$. The \emph{relevance} score is an 
exogeneous measure of how well an item answers a query, that is, it is the assumed system metric for the susbet selection task at hand.

\paragraph{Representativeness of an item.}

Let $\rep{\type}{i}{p}{\query}\in [-1,1]$ measure the
\emph{representativeness} of an item $i\in\inst_\query$ for $p$ and
query $\query$ along attribute $\type$.  Representativeness 
measures how well an item aligns with a user
$p$'s attribute $\type$ (e.g., if $i$ has similar hair texture to $p$
and $q$ refers to hair styles). We allow for representativeness to be
both positive and negative, to capture the idea that an item might be
a positive or negative representation of $p$, and that this polarity
might depend on $\query$ as well as the attribute $\type$.

There are many natural choices for the representativeness
function. For example, if items correspond to people, then a candidate
representativeness function could indicate whether the attribute is
the same for both $p$ and an item:
\[\rep{\type}{i}{p}{\query} = \I[\type(\{i\}) = \type(\{p\})].\]

One could also choose some more complex measure of the match of $i$ to
$p$ along $\type$. One can similarly define a notion of
representativeness for items that are not individuals, if individuals
find some of those items as being well-aligned with their identity
along $\type$.

This can express that ``similar" individuals along $a$ may make $p$ feel more included, even if similar values to $\type$ 
would not increase the diversity score for $A = \{\type\}$.  This captures the idea that
the diversity score measures an abstracted and simplified summary,
while the inclusion score affords a more fluid contextual
understanding of identities. 

An instance's set of items, their relevance, and their
representativeness together may be represented as: 
\[r_{\inst_\query} = \{ (i, \rel{i}{\query},
  \rep{\type}{i}{p}{\query}) | i \in \inst_\query, \query\}.\] We can
then define the inclusion of an instance as an aggregate statistic of
the set of items in the instance, their relevance to the query, and
the items' alignment or match to individual $p$ along $\type$:
\[\inclusion{\type}{\inst_\query}{p}{\query} = f(r_{\inst_\query}) \in
  [-1,1].\]

In the simplest case, each instance $x_q$ may contain only one item
(or one relevant item), in which case $f$ might simply
  report the representativeness of the single (relevant) item. In the
case where many items in an instance are relevant, $f$ might measure
the median representativeness of the high-relevance items in
$\inst_\query$, or the maximum representativeness of some item in the
instance. 

An inclusion score near $-1$ indicates $p$ finds the instance
stereotypical; this is similar to the notion of negative stereotypes
in representation \cite{cheryan2013stereotypical} or tokenism
\cite{tokenism}. A score near $1$ refers to $p$'s known attribute
$\type$ being well aligned in $\inst_\query$. A score near $0$
corresponds to $p$ finding few or no attribute alignments in
$\inst_\query$.

\paragraph{Set Inclusion.}
An instance giving rise to an inclusion score for $p$ along an
attribute $\type$ for query $\query$ naturally extends to scoring the inclusion of a set of instances.  The \textit{cumulative inclusion
  score} of a set $\Set_\query$ is a function of
$\inc{\type}{\inst_{\query}}{p}{\query}$ across the instances in the
set: $\inc{\type}{\Set_{\query}}{p}{\query} = g\left( \{\inc{\type}{\inst}{p }{\query} | \inst \in \Set_\query\}\right)$.  In this formulation, the inclusion score of an instance is comprised
of the representativeness and relevance of items within it, and the
inclusion score of a set is made up of the instances within the set.

\paragraph{Multiple Attribute Inclusion.}
Another type of cumulative inclusion score ranges over the set of
attributes known about $p$, capturing a holistic sense of inclusion
for $p$ rather than one according to a single attribute. Just as in
set inclusion, many natural definitions of multiple attribute inclusion
arise from defining a cumulative function $g(\cdot)$.  

Both
instance-based and attribute-based cumulative functions for Inclusion
can leverage social choice theory to return the final score, as detailed in the Social Choice section below.  For
example, in a Nash Welfare Inclusivity approach for Set Inclusion,
$g(\cdot)$ would return the geometric mean over $\inc{a}{x}{p}{q}$ for
$\inst \in \Set$.  In a Nash Welfare Inclusivity approach for Multiple
Attribute Inclusion, $g(\cdot)$ would return the geometric mean over
$\inc{a}{x}{p}{q}$ for $\type \in \Types$.

\subsection{Inclusion Metrics Discussion}

\paragraph{The relevance function $\rel{}{}$.}
We now reflect on the relevance function in the description of
inclusion above. We mention above that the relevance function measures
how well an item corresponds to a query string $q$.  The objective
function for many subset selection algorithms often measures exactly
such a quantity, independent of inclusion or diversity concerns,
though this may only be measured for an instance $\inst_{\query}$
rather than items in the instance.

However, the ground-truth relevance score of an instance or set of
instances with respect to some $q$ may never be measurable or even
directly defined, and for this reason some simpler proxies are often
used in place of a ground truth relevance score.  If one uses this
same proxy score function to define inclusion, this choice may
affect inclusion scores for certain parties more than others due to
unequal measurement error across the space of items and
instances.

\begin{figure*}
\centering
\small
\begin{tabular}{llrrlrrlrr
ccc}
& \multicolumn{3}{c}{\multirow{2}{*}{\includegraphics[width=.2\textwidth]{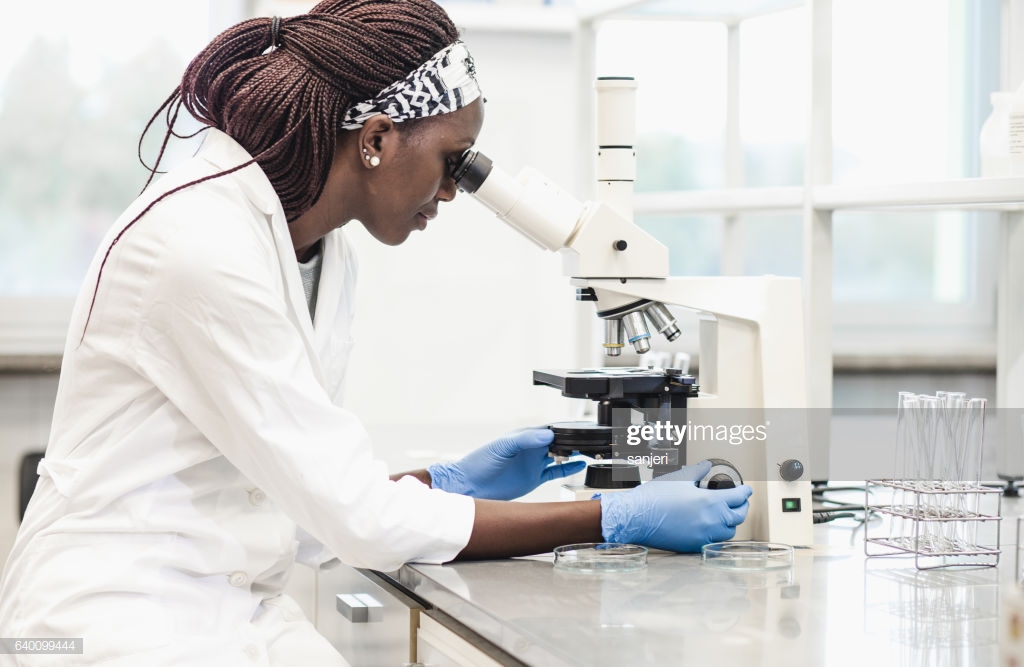}}}
& \multicolumn{3}{c}{\multirow{2}{*}{\includegraphics[trim={0 0.5cm 0 0},clip,width=.2\textwidth]{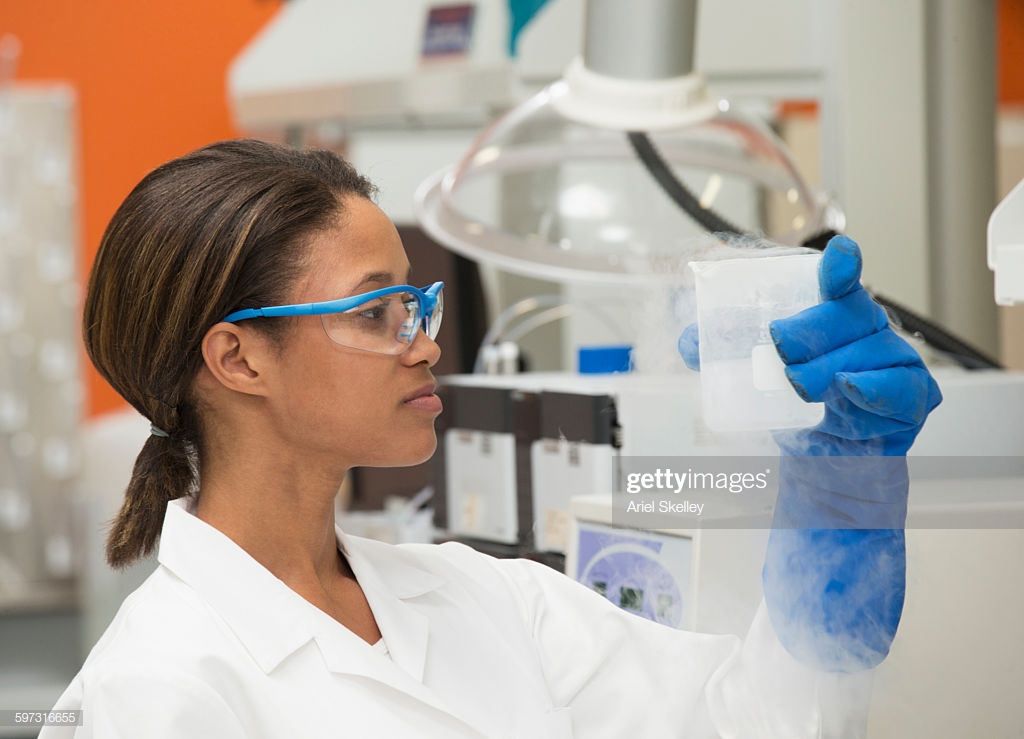}}}
& \multicolumn{3}{c}{\multirow{2}{*}{\includegraphics[width=.2\textwidth]{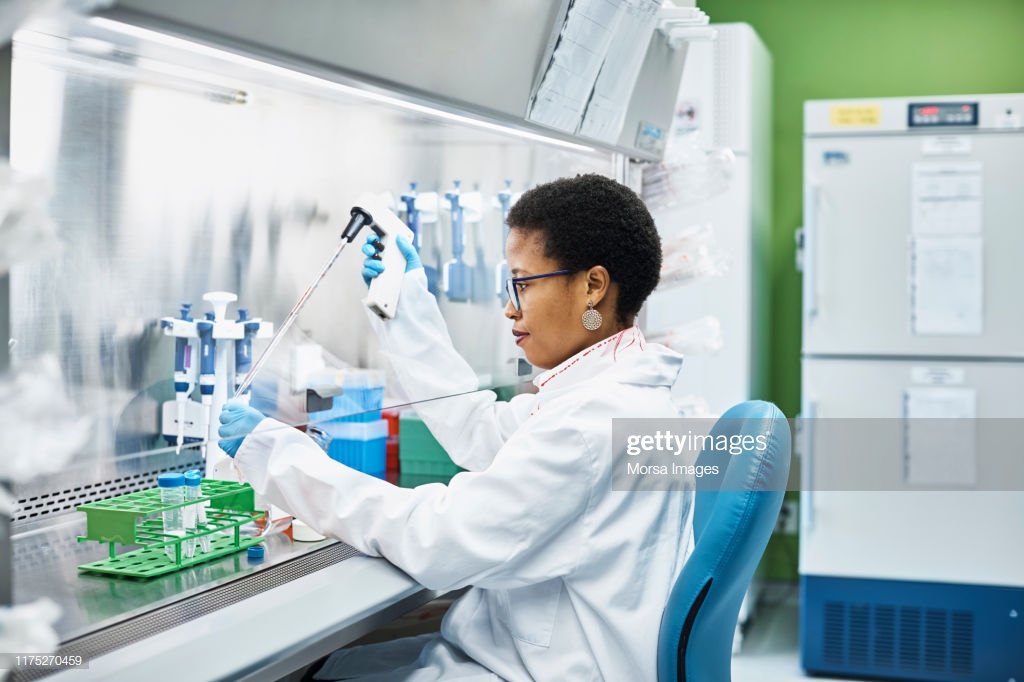}}} & 
\multicolumn{3}{l}{\large {\bf Given:}} \\
\\
&&&&&&&&&& \multicolumn{3}{l}{\normalsize{\textbf{Individual $p$:}}} \\
&&&&&&&&&& \multicolumn{3}{l}{gender:woman} \\ &&&&&&&&&& \multicolumn{3}{l}{skin:type\_6, age:70} \\
\\
&&&&&&&&&& \multicolumn{3}{l}{\normalsize{\textbf{Query $q$:} Scientist}} \\
{$\inc{\type}{x}{p}{q}$}
    &\multicolumn{3}{c}{\large{\bf $x_1$}} & 
    \multicolumn{3}{c}{\large{\bf $x_2$}} &
    \multicolumn{3}{c}{\large{\bf $x_3$}} \\
    \cline{1-10}

    \multicolumn{1}{|l|}{Inc$_{\text{gender:woman}}$}
    &  \hspace{.5em}woman & 1 & \multicolumn{1}{@{}l;{2pt/2pt}}{= 1.00}
    &  \hspace{.5em}woman & 1 & \multicolumn{1}{@{}l;{2pt/2pt}}{= 1.00}
    &  \hspace{.5em}woman & 1 & \multicolumn{1}{@{}l|}{= 1.00}
&\multicolumn{3}{c}{}\\    

    \multicolumn{1}{|l|}{Inc$_{\text{skin:type\_6}}$}
    & \hspace{.5em}type\_5 & $\frac{5}{6}$ & \multicolumn{1}{@{}l;{2pt/2pt}}{= 0.83}
    & \hspace{.5em}type\_4 & $\frac{4}{6}$ & \multicolumn{1}{@{}l;{2pt/2pt}}{= 0.67}
    & \hspace{.5em}type\_3 & $\frac{3}{6}$ & \multicolumn{1}{@{}l|}{= 0.50}
& \multicolumn{3}{c}{}\\

    \multicolumn{1}{|l|}{Inc$_{\text{age:70}}$}
    & \hspace{.5em}31 & $1-(\frac{70 - 31}{100})$ & \multicolumn{1}{@{}l;{2pt/2pt}}{= 0.61}
    & \hspace{.5em}23 & $1-(\frac{70 - 23}{100})$ & \multicolumn{1}{@{}l;{2pt/2pt}}{= 0.53}
    & \hspace{.5em}47 & $1-(\frac{70 - 47}{100})$ & \multicolumn{1}{@{}l|}{= 0.77}
    \\
    \hline 
    \multicolumn{1}{|l|}{\bf Cumulative} & \multicolumn{3}{c;{2pt/2pt}}{Multiple Attribute} & \multicolumn{3}{c;{2pt/2pt}}{Multiple Attribute}&\multicolumn{3}{c|}{Multiple Attribute}& \multicolumn{3}{c|}{Set (Pair)} \\
    \multicolumn{1}{|l|}{\bf Scores} &\multicolumn{3}{c;{2pt/2pt}}{{\large{\bf $x_1$}}} &
    \multicolumn{3}{c;{2pt/2pt}}{\large{\bf $x_2$}} &
    \multicolumn{3}{c|}{\large{\bf $x_3$}}& \multicolumn{1}{c@{}}{\large{\bf $\{x_1,x_2\}$}} & \multicolumn{1}{c@{}}{\large{\bf $\{x_1,x_3\}$}} & \multicolumn{1}{c|}{\large{\bf $\{x_2,x_3\}$}} \\\hline
    \multicolumn{1}{|l|}{Utilitarian} &&& \multicolumn{1}{c;{2pt/2pt}}{0.81} &&& \multicolumn{1}{c;{2pt/2pt}}{0.73} &&& \multicolumn{1}{c}{0.76}
    & \multicolumn{1}{|@{}c@{}}{0.77} & \multicolumn{1}{@{}c@{}}{\bf 0.79} & \multicolumn{1}{@{}c@{}|}{0.75}\\\cdashline{2-13}[2pt/2pt]
    \multicolumn{1}{|l|}{Egalitarian} &&& \multicolumn{1}{c;{2pt/2pt}}{0.61} &&& \multicolumn{1}{c;{2pt/2pt}}{0.53} &&& \multicolumn{1}{c}{0.50}
    & \multicolumn{1}{|@{}c@{}}{\bf 0.53} & \multicolumn{1}{@{}c@{}}{0.50} & \multicolumn{1}{@{}c@{}|}{0.50}\\\cdashline{2-13}[2pt/2pt]
    \multicolumn{1}{|l|}{Nash} &&& \multicolumn{1}{c;{2pt/2pt}}{0.79} &&& \multicolumn{1}{c;{2pt/2pt}}{0.71} &&& \multicolumn{1}{c}{0.73}
    & \multicolumn{1}{|@{}c@{}}{0.75} & \multicolumn{1}{@{}c@{}}{\bf 0.76} & \multicolumn{1}{@{}c@{}|}{0.72}\\\hline
    
\end{tabular}%
\vspace{-0.5em}
\caption{Worked example of Inclusion scores for attributes of each instance $x_q$, given a user $p$ and a query $q$. Below each image are the associated attributes (left) and the Inclusion scores for $p$ on this attribute (right). In this example, we must select two images out of the three. Three different methods for aggregating the inclusion scores for attributes are illustrated. The first, motivated by utilitarianism, takes the average inclusion score for the image pair. The highest-scoring pair is then images x$_1$ and x$_3$ images.
The second, motivated by egalitarianism, takes the minimum inclusion score of the pair. The highest-scoring pair is then images x$_1$ and x$_2$. Finally, Nash inclusivity chooses the pair with the highest geometric mean, in this case the same images as in utilitarianism, x$_1$ and x$_3$.}\vspace{-1em}
\label{alt_scientist_example}
\end{figure*}

\section{Comparing Subset Inclusivity: Approaches from Social Choice Theory}\label{sec:econ} 

We have defined Diversity and Inclusion criteria for single attributes in single instances, and have briefly discussed how these can be extended to sets of instances or to sets of attributes.
Extending to such sets requires a {\it cumulation} mechanism, which produces a single score from a set of scores. Here, we can build from social choice theory, which has well-developed mechanisms for determining a final score from a set of scored items based on the ethical goals defined for a system. For example, an \textit{egalitarian} mechanism \cite{rawls1974some} can be used to favor under-served individuals that share an attribute. A \textit{utilitarian} mechanism \cite{mill2016utilitarianism} can be used to treat all attributes as equally important, producing an arithmetic average over items. Such methods may also be used to compare scores across sets. We detail three such relevant mechanisms for subset scoring below, and illustrate these concepts using scores in Figure 2.

\textbf{Egalitarian (maximin) inclusivity}. Set $X_1$ may be said to be more inclusive than set $X_2$ if the lowest inclusion score in $X_1$ is higher than the lowest inclusion score in $X_2$, i.e., \[
  \min_i(\Set_{1i})>\min_i(\Set_{2i}).\] If $\min_i(\Set_{1i}) = \min_i(\Set_{2i})$, then repeat for the second lowest scores, third, and so on. If the two mechanisms are equal, we are indifferent between $\Set_1$ and $\Set_2$. 
  

\textbf{Utilitarian inclusivity}: This corresponds to an arithmetic average over the inclusion scores for all items in the set, where a set $\Set_1$ is
more inclusive than $\Set_2$ if the average of its inclusion metric scores is greater.
\[
\frac{1}{n}\sum\limits_{i}{\Set_{2i}} < \frac{1}{n}\sum\limits_{i}{\Set_{1i}}  
\].

\textbf{Nash inclusivity}: This corresponds to the geometric mean over the inclusion scores for all items in the set. Set $\Set_1$ is more inclusive than $\Set_2$ if the product of its inclusion metric scores is greater, i.e.,
\[\sqrt[\leftroot{-2}\uproot{2}n]{\prod\limits_{i}{\Set_{2i}}} < \sqrt[\leftroot{-2}\uproot{2}n]{\prod\limits_{i}{\Set_{1i}}}\]
Nash inclusivity can be seen as a mix of utilitarian and egalitarian, as it monotonically increases with both of these measures \cite{caragiannis2019unreasonable}. 

\section{Metrics In Practice}

We assume that $Z$ is a set of instances relevant to the domain of
interest $\query$, such that instances within each selected subset
$\Set_\query$ are relevant according to $\rel{}{}$, where a score of
1.0 means that an instance is relevant to the query.

\paragraph{Prompt Polarity.} When applying Diversity and Inclusion metrics in a domain where the query is not only neutral, but may also be negative (e.g., ``jerks''), it is necessary to incorporate a $\text{polarity}(q)$ value into the score to tease out the `negative' meaning and values of the inclusion
score, as may be provided by a sentiment model. For example:
\[\rep{\type}{i}{p}{\query} = \I[\type(\{i\}) = \type(\{p\})] * \lambda \text{polarity}(\query).\]

\paragraph{Stereotyping.}
  Note that the $X_q$ subset for a given $<$p, q$>$ pair can increase the diversity score by producing diverse stereotypes\footnote{Examples of stereotypes intentionally omitted throughout paper in order to minimize further stereotype propagation.} unless $p$ and $q$ are well defined.  
The domain of relevance $q$ is crucial for understanding whether a set of
results might stereotype by a particular attribute.  For
example, if $q$ is ``work clothing'', and the set $X$ contains only pink womens' workwear but a variety of colors for mens'
workwear, this set could be said to uphold the stereotype about women
and their color preferences, even if the set is diverse and inclusive for a man.  On the other hand, if $q$ is ``pink womens' work clothing'', the same set of womens'
clothing reflects the query and domain, while in the former case the
results overconcentrate a specific color in the results relevant to
women. \textit{Stereotyping} here refers to homogeneity across results for attribute $\type \in \Types$.

The person perceiving a set of results $X$ is obviously the arbiter of whether the results stereotype them. Suppose the
person searching for clothing in the previous example is a
woman. If she likes pink workwear, she might feel as though the
instances of womens' workwear being pink suits her goals and needs; if
she does not particularly like pink, even if a majority of women
generally like pink, the results of a search containing only pink
womens' clothing does not meet her goals, but does reinforce a
standard assumption about womens' clothing.

\paragraph{Intersectionality}
Crossing demographics-based $\Types$ such as those based on Gender and Race yields intersectional $\Types$ that can be
applied in the same manner as unitary $\Types$.
Without accounting for intersectionality, it is possible for a set of instances to receive
high diversity and inclusion scores without reflecting the unique
characteristics of the individual.  For example, if a black woman is searching for movie
recommendations, and the set returned is half movies starring black
men and half movies starring white women, the selection may be diverse and aligned somewhat with her social identities while still creating a sense of exclusion.

\paragraph{Inclusion within Instances.}
The focus of the family of inclusion metrics introduced in this paper is inclusion towards the individual
presented with the set.  Another aspect of inclusion concerns the
individuals represented in the instances.  For
example, if $\Set_\query$ contains people of different
ethnicities, all stereotyped except for the one that authentically
represents the ethnicity of the individual, the proposed metrics will
not capture this effect. It may be desirable to apply the Inclusion
metric not only to the individual creating the query, but also to
those who may be represented. 
 
\subsection{Worked Example}

We begin with the context and person creating the query. The person may be seeking a selection of stock images to use for a presentation to an unknown-to-them audience. The person has a token $p$ in the system where they permit information to be stored, such as their gender and hair color. Assume a specific $p$: $\{\text{gender:female, skin:6, age:70, hair:shortgrey}\}$.\footnote{$\text{skin:6}$ refers to Fitzpatrick Skin Type 6 \cite{Fitz}.}  A generalization is a list of attributes most at risk for disproportionately unfair experiences, without requiring correspondence to a specific individual.  $\inclusion{a}{x}{p}{q}$ scores are shown in Figure 2. Each image $x_q$ has one item $i$, and for simplicity we assume the given relevance score for all images $\rel{i}{q} = 1$.\footnote{That is, all images are equally relevant to the query.}  The Inclusion score is then:
 {\footnotesize
\begin{align*}\inclusion{\type}{\inst_\query}{p}{\query} &= f(r_{\inst_\query}) \\
&=
\rel{i}{q} * \rep{a}{i}{p}{q} \\
&= \rep{a}{i}{p}{q}
\end{align*}}%
\vspace{-1.25em}

Inclusion is here equal to the $\text{representativeness}$ score for each group type ($\text{skin, age, hair}$). Basic instantiations of the $\rep{a}{i}{p}{q}$ metric may be measures of distance or match:\vspace{-0.25em}
{\footnotesize
\begin{alignat*}{3}
&\rep{gender:p}{i}{p}{\text{scientist}} = && \quad i_{gender} \equiv p_{gender}\\
&\rep{skin:p}{i}{p}{\text{scientist}} = && \quad {MAX}_{\mathit{skin}} - \displaystyle\frac{\dist(i_{\mathit{skin}}, p_{\mathit{skin}})}{{MAX}_{\mathit{skin}}} \\ 
&\rep{age:p}{i}{p}{\text{scientist}} = && \quad  \displaystyle\frac{\dist(\mathit{age}_i, \mathit{age}_p)}{{MAX}_{\mathit{age}}} \\
&\rep{hair:p}{i}{p}{\text{scientist}} = && \quad  \mathit{texture}_i \equiv  \mathit{texture}_p \vee \qquad \qquad \\\vspace{-1em}
&                       &&  \quad \mathit{length}_i 
\equiv \mathit{length}_p \vee \qquad \qquad \\\vspace{-1em}
& && \quad \mathit{color}_i \equiv \mathit{color}_p \vee \qquad \qquad \\\vspace{-1.25em}
& &&  \quad \mathit{style}_i \equiv \mathit{style}_p \qquad \qquad
\end{alignat*}
}%

 Figure 2 details inclusion scores for a set of images $X_{\query}$ given the person $p$ described above. Applying the Diversity criteria above, with Presence scored by an indicator function, each image $x$ has a Diversity score of 0, because each attribute has only one form in each image (e.g., a single person is present).  The image set $X_q$ is also not Gender Diverse.



%

%
%

\section{Image Set Perception Study}

\subsection{Overview}
To evaluate the viability of our proposed metrics, we conducted surveys on Amazon's Mechanical Turk platform, asking respondents to compare the relative diversity and inclusiveness of sets of images with respect to \textit{gender} and \textit{skin tone}. 

To do this, we curated several stock image sets containing people depicting specific occupations, listed in Table \ref{tab:occupations}. These sets were designed to be diverse and/or inclusive as outlined in this paper. Specifically, we curated four sets of images: a set that was diverse but not inclusive (D+I-), inclusive but not diverse (D-I+), both inclusive and diverse (D+I+), and neither inclusive nor diverse (D-I-).

Respondents were presented with pairs of image sets from a given occupation and asked to select which was more inclusive or diverse with respect to a specified demographic---gender or skin tone---with an option to indicate that both were approximately the same. At the end of the survey, we also collected information on rater age and gender\footnote{Our interface also allowed us to collect genders beyond the man/woman binary. However, due to the small sample size, they are excluded from our analysis.}. We scored image sets by simply calculating the percentage of all comparisons where the image set ``won'' (i.e. was selected as the more diverse or inclusive set).

\vspace{-0.5em}
\begin{table}[ht]
\caption{Occupations in study}
\centering
\vspace{-1em}
\begin{tabular}{|cccc|}
\hline
computer programmer & scientist & doctor & nurse\\
salesperson & janitor & lawyer & dancer\\
\hline
\end{tabular}
\label{tab:occupations}
\end{table}

\vspace{-1em}
\subsection{Results}
\begin{figure}
    \centering
    \includegraphics[height=6cm]{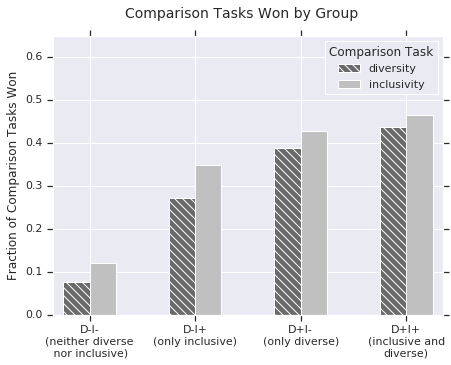}
    \caption{Fraction of comparison tasks ``won'' for each image group and task.}
    \label{fig:comparison_tasks_by_group}
\end{figure}

As shown in figure \ref{fig:comparison_tasks_by_group}, we found that
aggregating across occupations, D+I+ image sets had the highest
average scores for both the diversity and inclusion comparison tasks,
with D+I+ sets receiving higher diversity and inclusion ratings than
the other three conditions (D+I-, D-I+, and D-I-). D-I- sets received
the lowest diversity and inclusion ratings. This suggests, perhaps
unsurprisingly, that there is some overlap in the concepts of
diversity and inclusion: inclusivity
adds to the perception of diversity, and vice versa.

Although there is overlap in the perception of the two concepts, our
results also suggest that respondents differentiated between our metrics of inclusivity and
diversity. 
Specifically, D-I+ stimuli were labeled as more inclusive than
diverse, aligning with the intended diversity and inclusion of the
sets.  Interestingly, D+I- stimuli were {\it also} labeled as more
inclusive than diverse, although the gap between inclusion and
diversity ratings is smaller. 
These results indicate that respondents perceive sets with more diversity and inclusion over a baseline as more {\it inclusive} than diverse.  

When split by users' self-identified gender, men tended to
rate D+I- conditions as more inclusive than diverse, while women
tended to rate these conditions as equally inclusive and
diverse. Female respondents also found the D-I+ sets substantially
more inclusive than diverse, with much less of a difference between
diversity and inclusion scores for the remainder of the sets.  This
discrepancy underscores the relevancy of the user: the
identity of the respondents impacts perceptions of
diversity and inclusion in image sets.

\subsection{Data Quality}

We screened for low-quality responses using three approaches: duplicate ``confirmation questions'', the use of free-response fields on a multiple-choice question, and reCAPTCHA. First, each set of comparisons contained two ``confirmation questions'', which were simply duplicates of earlier questions with images shuffled and the comparison presented in reverse order. Second, while the survey had only three available options (``Set A'', ``Set B'', and ``Same''), respondents were given a free-response answer box to type their answer. 
This allowed us to filter for automated responses, as we found that a small fraction of the responses were nonsensical (e.g. ``No'', or ``Very good''). Finally, respondents had to fill out a reCAPTCHA form before submitting. 
Answers with a reCAPTCHA score below $0.5$, those whose confirmation questions did not agree, and free-response answers that could not be resolved into a valid response were removed. 
After filtering, we had 491 valid responses, which contained comparisons between all image sets for each occupation.
\section{Discussion}

We have distinguished between notions of \textit{diversity} and \textit{inclusion} and detailed how they may be formalized, applied to the general problem of scoring instances or sets.  This may be useful in subset selection problems that seek to reflect individuals with attributes that are disproportionately marginalized, such as when selecting images of people in a stock photo selection task. Our worked example demonstrates how social choice theory can be applied to compare diversity and inclusion scores across different sets. 

\begin{acks}
Thank you to Andrew Zaldivar, Ben Packer, and Tulsee Doshi for the insightful discussions and suggestions.
\end{acks}
\bibliographystyle{ACM-Reference-Format}
\bibliography{camera_ready_arxiv}


\begin{thebibliography}{37}


\ifx \showCODEN    \undefined \def \showCODEN     #1{\unskip}     \fi
\ifx \showDOI      \undefined \def \showDOI       #1{#1}\fi
\ifx \showISBNx    \undefined \def \showISBNx     #1{\unskip}     \fi
\ifx \showISBNxiii \undefined \def \showISBNxiii  #1{\unskip}     \fi
\ifx \showISSN     \undefined \def \showISSN      #1{\unskip}     \fi
\ifx \showLCCN     \undefined \def \showLCCN      #1{\unskip}     \fi
\ifx \shownote     \undefined \def \shownote      #1{#1}          \fi
\ifx \showarticletitle \undefined \def \showarticletitle #1{#1}   \fi
\ifx \showURL      \undefined \def \showURL       {\relax}        \fi
\providecommand\bibfield[2]{#2}
\providecommand\bibinfo[2]{#2}
\providecommand\natexlab[1]{#1}
\providecommand\showeprint[2][]{arXiv:#2}

\bibitem[\protect\citeauthoryear{Anari, Gharan, and Rezaei}{Anari
  et~al\mbox{.}}{2016}]%
        {23}
\bibfield{author}{\bibinfo{person}{Nima Anari}, \bibinfo{person}{Shayan~Oveis
  Gharan}, {and} \bibinfo{person}{Alireza Rezaei}.}
  \bibinfo{year}{2016}\natexlab{}.
\newblock \showarticletitle{Monte Carlo Markov chain algorithms for sampling
  strongly Rayleigh distributions and determinantal point processes}. In
  \bibinfo{booktitle}{\emph{Conference on Learning Theory}}.
  \bibinfo{pages}{103--115}.
\newblock


\bibitem[\protect\citeauthoryear{Asudeh, Jin, and Jagadish}{Asudeh
  et~al\mbox{.}}{2019}]%
        {assessing}
\bibfield{author}{\bibinfo{person}{Abolfazl Asudeh}, \bibinfo{person}{Zhongjun
  Jin}, {and} \bibinfo{person}{HV Jagadish}.} \bibinfo{year}{2019}\natexlab{}.
\newblock \showarticletitle{Assessing and remedying coverage for a given
  dataset}. In \bibinfo{booktitle}{\emph{2019 IEEE 35th International
  Conference on Data Engineering (ICDE)}}. IEEE, \bibinfo{pages}{554--565}.
\newblock


\bibitem[\protect\citeauthoryear{Barocas and Selbst}{Barocas and
  Selbst}{2016}]%
        {barocas2016big}
\bibfield{author}{\bibinfo{person}{Solon Barocas} {and}
  \bibinfo{person}{Andrew~D Selbst}.} \bibinfo{year}{2016}\natexlab{}.
\newblock \showarticletitle{Big data's disparate impact}.
\newblock \bibinfo{journal}{\emph{Calif. L. Rev.}}  \bibinfo{volume}{104}
  (\bibinfo{year}{2016}), \bibinfo{pages}{671}.
\newblock


\bibitem[\protect\citeauthoryear{Baselga, Jim{\'e}nez-Valverde, and
  Niccolini}{Baselga et~al\mbox{.}}{2007}]%
        {bio1}
\bibfield{author}{\bibinfo{person}{Andr{\'e}s Baselga},
  \bibinfo{person}{Alberto Jim{\'e}nez-Valverde}, {and} \bibinfo{person}{Gilles
  Niccolini}.} \bibinfo{year}{2007}\natexlab{}.
\newblock \showarticletitle{A multiple-site similarity measure independent of
  richness}.
\newblock \bibinfo{journal}{\emph{Biology Letters}} \bibinfo{volume}{3},
  \bibinfo{number}{6} (\bibinfo{year}{2007}), \bibinfo{pages}{642--645}.
\newblock


\bibitem[\protect\citeauthoryear{Bell and Hartmann}{Bell and Hartmann}{2007}]%
        {bell2007diversity}
\bibfield{author}{\bibinfo{person}{Joyce~M Bell} {and} \bibinfo{person}{Douglas
  Hartmann}.} \bibinfo{year}{2007}\natexlab{}.
\newblock \showarticletitle{Diversity in everyday discourse: The cultural
  ambiguities and consequences of “happy talk”}.
\newblock \bibinfo{journal}{\emph{American Sociological Review}}
  \bibinfo{volume}{72}, \bibinfo{number}{6} (\bibinfo{year}{2007}),
  \bibinfo{pages}{895--914}.
\newblock


\bibitem[\protect\citeauthoryear{Berrey}{Berrey}{2015}]%
        {berrey2015enigma}
\bibfield{author}{\bibinfo{person}{Ellen Berrey}.}
  \bibinfo{year}{2015}\natexlab{}.
\newblock \bibinfo{booktitle}{\emph{The Enigma of Diversity: The Language of
  Race and the Limits of Racial Justice}}.
\newblock \bibinfo{publisher}{University of Chicago Press}.
\newblock


\bibitem[\protect\citeauthoryear{Caragiannis, Kurokawa, Moulin, Procaccia,
  Shah, and Wang}{Caragiannis et~al\mbox{.}}{2019}]%
        {caragiannis2019unreasonable}
\bibfield{author}{\bibinfo{person}{Ioannis Caragiannis}, \bibinfo{person}{David
  Kurokawa}, \bibinfo{person}{Herv{\'e} Moulin}, \bibinfo{person}{Ariel~D
  Procaccia}, \bibinfo{person}{Nisarg Shah}, {and} \bibinfo{person}{Junxing
  Wang}.} \bibinfo{year}{2019}\natexlab{}.
\newblock \showarticletitle{The unreasonable fairness of maximum Nash welfare}.
\newblock \bibinfo{journal}{\emph{ACM Transactions on Economics and Computation
  (TEAC)}} \bibinfo{volume}{7}, \bibinfo{number}{3} (\bibinfo{year}{2019}),
  \bibinfo{pages}{12}.
\newblock


\bibitem[\protect\citeauthoryear{Celis, Deshpande, Kathuria, and Vishnoi}{Celis
  et~al\mbox{.}}{2016}]%
        {fair_diverse}
\bibfield{author}{\bibinfo{person}{L~Elisa Celis}, \bibinfo{person}{Amit
  Deshpande}, \bibinfo{person}{Tarun Kathuria}, {and}
  \bibinfo{person}{Nisheeth~K Vishnoi}.} \bibinfo{year}{2016}\natexlab{}.
\newblock \showarticletitle{How to be fair and diverse?}
\newblock \bibinfo{journal}{\emph{arXiv preprint arXiv:1610.07183}}
  (\bibinfo{year}{2016}).
\newblock


\bibitem[\protect\citeauthoryear{Celis, Straszak, and Vishnoi}{Celis
  et~al\mbox{.}}{2017}]%
        {CelisEtAl17}
\bibfield{author}{\bibinfo{person}{L.~Elisa Celis}, \bibinfo{person}{Damian
  Straszak}, {and} \bibinfo{person}{Nisheeth~K. Vishnoi}.}
  \bibinfo{year}{2017}\natexlab{}.
\newblock \showarticletitle{Ranking with Fairness Constraints}.
\newblock \bibinfo{journal}{\emph{CoRR}}  \bibinfo{volume}{abs/1704.06840}
  (\bibinfo{year}{2017}).
\newblock
\showeprint[arxiv]{1704.06840}
\urldef\tempurl%
\url{http://arxiv.org/abs/1704.06840}
\showURL{%
\tempurl}


\bibitem[\protect\citeauthoryear{Cheryan, Plaut, Handron, and Hudson}{Cheryan
  et~al\mbox{.}}{2013}]%
        {cheryan2013stereotypical}
\bibfield{author}{\bibinfo{person}{Sapna Cheryan}, \bibinfo{person}{Victoria~C
  Plaut}, \bibinfo{person}{Caitlin Handron}, {and} \bibinfo{person}{Lauren
  Hudson}.} \bibinfo{year}{2013}\natexlab{}.
\newblock \showarticletitle{The stereotypical computer scientist: Gendered
  media representations as a barrier to inclusion for women}.
\newblock \bibinfo{journal}{\emph{Sex roles}} \bibinfo{volume}{69},
  \bibinfo{number}{1-2} (\bibinfo{year}{2013}), \bibinfo{pages}{58--71}.
\newblock


\bibitem[\protect\citeauthoryear{Deshpande and Rademacher}{Deshpande and
  Rademacher}{2010}]%
        {24}
\bibfield{author}{\bibinfo{person}{Amit Deshpande} {and} \bibinfo{person}{Luis
  Rademacher}.} \bibinfo{year}{2010}\natexlab{}.
\newblock \showarticletitle{Efficient volume sampling for row/column subset
  selection}. In \bibinfo{booktitle}{\emph{2010 IEEE 51st Annual Symposium on
  Foundations of Computer Science}}. IEEE, \bibinfo{pages}{329--338}.
\newblock


\bibitem[\protect\citeauthoryear{Dobbin and Kalev}{Dobbin and Kalev}{2016}]%
        {dobbin2016diversity}
\bibfield{author}{\bibinfo{person}{Frank Dobbin} {and}
  \bibinfo{person}{Alexandra Kalev}.} \bibinfo{year}{2016}\natexlab{}.
\newblock \showarticletitle{Why Diversity Programs Fail and What Works Better}.
\newblock \bibinfo{journal}{\emph{Harvard Business Review}}
  \bibinfo{volume}{94}, \bibinfo{number}{7-8} (\bibinfo{year}{2016}),
  \bibinfo{pages}{52--60}.
\newblock


\bibitem[\protect\citeauthoryear{Drosou, Jagadish, Pitoura, and
  Stoyanovich}{Drosou et~al\mbox{.}}{2017}]%
        {fairness1}
\bibfield{author}{\bibinfo{person}{Marina Drosou}, \bibinfo{person}{HV
  Jagadish}, \bibinfo{person}{Evaggelia Pitoura}, {and} \bibinfo{person}{Julia
  Stoyanovich}.} \bibinfo{year}{2017}\natexlab{}.
\newblock \showarticletitle{Diversity in big data: A review}.
\newblock \bibinfo{journal}{\emph{Big data}} \bibinfo{volume}{5},
  \bibinfo{number}{2} (\bibinfo{year}{2017}), \bibinfo{pages}{73--84}.
\newblock


\bibitem[\protect\citeauthoryear{Dwork, Hardt, Pitassi, Reingold, and
  Zemel}{Dwork et~al\mbox{.}}{2012}]%
        {dwork2012fairness}
\bibfield{author}{\bibinfo{person}{Cynthia Dwork}, \bibinfo{person}{Moritz
  Hardt}, \bibinfo{person}{Toniann Pitassi}, \bibinfo{person}{Omer Reingold},
  {and} \bibinfo{person}{Richard Zemel}.} \bibinfo{year}{2012}\natexlab{}.
\newblock \showarticletitle{Fairness through awareness}. In
  \bibinfo{booktitle}{\emph{Proceedings of the 3rd innovations in theoretical
  computer science conference}}. ACM, \bibinfo{pages}{214--226}.
\newblock


\bibitem[\protect\citeauthoryear{Embrick}{Embrick}{2011}]%
        {embrick2011diversity}
\bibfield{author}{\bibinfo{person}{David~G Embrick}.}
  \bibinfo{year}{2011}\natexlab{}.
\newblock \showarticletitle{The diversity ideology in the business world: A new
  oppression for a new age}.
\newblock \bibinfo{journal}{\emph{Critical sociology}} \bibinfo{volume}{37},
  \bibinfo{number}{5} (\bibinfo{year}{2011}), \bibinfo{pages}{541--556}.
\newblock


\bibitem[\protect\citeauthoryear{Fitzpatrick}{Fitzpatrick}{1988}]%
        {Fitz}
\bibfield{author}{\bibinfo{person}{Thomas~B. Fitzpatrick}.}
  \bibinfo{year}{1988}\natexlab{}.
\newblock \showarticletitle{{The Validity and Practicality of Sun-Reactive Skin
  Types I Through VI}}.
\newblock \bibinfo{journal}{\emph{JAMA Dermatology}} \bibinfo{volume}{124},
  \bibinfo{number}{6} (\bibinfo{date}{06} \bibinfo{year}{1988}),
  \bibinfo{pages}{869--871}.
\newblock
\showISSN{2168-6068}
\urldef\tempurl%
\url{https://doi.org/10.1001/archderm.1988.01670060015008}
\showDOI{\tempurl}


\bibitem[\protect\citeauthoryear{Gong, Chao, Grauman, and Sha}{Gong
  et~al\mbox{.}}{2014}]%
        {13}
\bibfield{author}{\bibinfo{person}{Boqing Gong}, \bibinfo{person}{Wei-Lun
  Chao}, \bibinfo{person}{Kristen Grauman}, {and} \bibinfo{person}{Fei Sha}.}
  \bibinfo{year}{2014}\natexlab{}.
\newblock \showarticletitle{Diverse sequential subset selection for supervised
  video summarization}. In \bibinfo{booktitle}{\emph{Advances in Neural
  Information Processing Systems}}. \bibinfo{pages}{2069--2077}.
\newblock


\bibitem[\protect\citeauthoryear{Jost et~al\mbox{.}}{Jost
  et~al\mbox{.}}{2009}]%
        {bio2}
\bibfield{author}{\bibinfo{person}{Lou Jost} {et~al\mbox{.}}}
  \bibinfo{year}{2009}\natexlab{}.
\newblock \showarticletitle{Mismeasuring biological diversity: response to
  Hoffmann and Hoffmann (2008)}.
\newblock \bibinfo{journal}{\emph{Ecological Economics}} \bibinfo{volume}{68},
  \bibinfo{number}{4} (\bibinfo{year}{2009}), \bibinfo{pages}{925--928}.
\newblock


\bibitem[\protect\citeauthoryear{Kalev, Dobbin, and Kelly}{Kalev
  et~al\mbox{.}}{2006}]%
        {kalev2006best}
\bibfield{author}{\bibinfo{person}{Alexandra Kalev}, \bibinfo{person}{Frank
  Dobbin}, {and} \bibinfo{person}{Erin Kelly}.}
  \bibinfo{year}{2006}\natexlab{}.
\newblock \showarticletitle{Best practices or best guesses? Assessing the
  efficacy of corporate affirmative action and diversity policies}.
\newblock \bibinfo{journal}{\emph{American sociological review}}
  \bibinfo{volume}{71}, \bibinfo{number}{4} (\bibinfo{year}{2006}),
  \bibinfo{pages}{589--617}.
\newblock


\bibitem[\protect\citeauthoryear{Kulesza, Taskar, et~al\mbox{.}}{Kulesza
  et~al\mbox{.}}{2012}]%
        {7}
\bibfield{author}{\bibinfo{person}{Alex Kulesza}, \bibinfo{person}{Ben Taskar},
  {et~al\mbox{.}}} \bibinfo{year}{2012}\natexlab{}.
\newblock \showarticletitle{Determinantal point processes for machine
  learning}.
\newblock \bibinfo{journal}{\emph{Foundations and Trends{\textregistered} in
  Machine Learning}} \bibinfo{volume}{5}, \bibinfo{number}{2--3}
  (\bibinfo{year}{2012}), \bibinfo{pages}{123--286}.
\newblock


\bibitem[\protect\citeauthoryear{Legendre, Borcard, and Peres-Neto}{Legendre
  et~al\mbox{.}}{2008}]%
        {ecology3}
\bibfield{author}{\bibinfo{person}{Pierre Legendre}, \bibinfo{person}{Daniel
  Borcard}, {and} \bibinfo{person}{Pedro~R Peres-Neto}.}
  \bibinfo{year}{2008}\natexlab{}.
\newblock \showarticletitle{Analyzing or explaining beta diversity? Comment}.
\newblock \bibinfo{journal}{\emph{Ecology}} \bibinfo{volume}{89},
  \bibinfo{number}{11} (\bibinfo{year}{2008}), \bibinfo{pages}{3238--3244}.
\newblock


\bibitem[\protect\citeauthoryear{Lin and Bilmes}{Lin and Bilmes}{2012}]%
        {14}
\bibfield{author}{\bibinfo{person}{Hui Lin} {and} \bibinfo{person}{Jeff~A
  Bilmes}.} \bibinfo{year}{2012}\natexlab{}.
\newblock \showarticletitle{Learning mixtures of submodular shells with
  application to document summarization}.
\newblock \bibinfo{journal}{\emph{arXiv preprint arXiv:1210.4871}}
  (\bibinfo{year}{2012}).
\newblock


\bibitem[\protect\citeauthoryear{MacArthur}{MacArthur}{1965}]%
        {bio3}
\bibfield{author}{\bibinfo{person}{Robert~H MacArthur}.}
  \bibinfo{year}{1965}\natexlab{}.
\newblock \showarticletitle{Patterns of species diversity}.
\newblock \bibinfo{journal}{\emph{Biological reviews}} \bibinfo{volume}{40},
  \bibinfo{number}{4} (\bibinfo{year}{1965}), \bibinfo{pages}{510--533}.
\newblock


\bibitem[\protect\citeauthoryear{Mill}{Mill}{2016}]%
        {mill2016utilitarianism}
\bibfield{author}{\bibinfo{person}{John~Stuart Mill}.}
  \bibinfo{year}{2016}\natexlab{}.
\newblock \showarticletitle{Utilitarianism}.
\newblock In \bibinfo{booktitle}{\emph{Seven masterpieces of philosophy}}.
  \bibinfo{publisher}{Routledge}, \bibinfo{pages}{337--383}.
\newblock


\bibitem[\protect\citeauthoryear{Mor-Barak and Cherin}{Mor-Barak and
  Cherin}{1998}]%
        {mor1998tool}
\bibfield{author}{\bibinfo{person}{Michal~E Mor-Barak} {and}
  \bibinfo{person}{David~A Cherin}.} \bibinfo{year}{1998}\natexlab{}.
\newblock \showarticletitle{A tool to expand organizational understanding of
  workforce diversity: Exploring a measure of inclusion-exclusion}.
\newblock \bibinfo{journal}{\emph{Administration in Social Work}}
  \bibinfo{volume}{22}, \bibinfo{number}{1} (\bibinfo{year}{1998}),
  \bibinfo{pages}{47--64}.
\newblock


\bibitem[\protect\citeauthoryear{Paradiso}{Paradiso}{2017}]%
        {HRThing}
\bibfield{author}{\bibinfo{person}{Anthony Paradiso}.}
  \bibinfo{year}{2017}\natexlab{}.
\newblock \showarticletitle{Diversity is Being Asked to the Party. Inclusion is
  Being Asked to Dance. \#SHRMDIV}.
\newblock \bibinfo{journal}{\emph{The Society for Human Resource Management
  (SHRM) Blog}} (\bibinfo{year}{2017}).
\newblock
\urldef\tempurl%
\url{https://blog.shrm.org/blog/diversity-is-being-asked-to-the-party-inclusion-is-being-asked-to-dance-shr}
\showURL{%
\tempurl}


\bibitem[\protect\citeauthoryear{Pelled, Ledford, and Mohrman}{Pelled
  et~al\mbox{.}}{1999}]%
        {hope1999demographic}
\bibfield{author}{\bibinfo{person}{Lisa~H. Pelled}, \bibinfo{person}{Gerald~E
  Ledford, Jr}, {and} \bibinfo{person}{Susan~A. Mohrman}.}
  \bibinfo{year}{1999}\natexlab{}.
\newblock \showarticletitle{Demographic dissimilarity and workplace inclusion}.
\newblock \bibinfo{journal}{\emph{Journal of Management studies}}
  \bibinfo{volume}{36}, \bibinfo{number}{7} (\bibinfo{year}{1999}),
  \bibinfo{pages}{1013--1031}.
\newblock


\bibitem[\protect\citeauthoryear{Rawls}{Rawls}{1974}]%
        {rawls1974some}
\bibfield{author}{\bibinfo{person}{John Rawls}.}
  \bibinfo{year}{1974}\natexlab{}.
\newblock \showarticletitle{Some reasons for the maximin criterion}.
\newblock \bibinfo{journal}{\emph{The American Economic Review}}
  \bibinfo{volume}{64}, \bibinfo{number}{2} (\bibinfo{year}{1974}),
  \bibinfo{pages}{141--146}.
\newblock


\bibitem[\protect\citeauthoryear{Roberson}{Roberson}{2006}]%
        {inclusion2}
\bibfield{author}{\bibinfo{person}{Quinetta~M Roberson}.}
  \bibinfo{year}{2006}\natexlab{}.
\newblock \showarticletitle{Disentangling the meanings of diversity and
  inclusion in organizations}.
\newblock \bibinfo{journal}{\emph{Group \& Organization Management}}
  \bibinfo{volume}{31}, \bibinfo{number}{2} (\bibinfo{year}{2006}),
  \bibinfo{pages}{212--236}.
\newblock


\bibitem[\protect\citeauthoryear{Samadi, Tantipongpipat, Morgenstern, Singh,
  and Vempala}{Samadi et~al\mbox{.}}{2018}]%
        {samadi2018price}
\bibfield{author}{\bibinfo{person}{Samira Samadi}, \bibinfo{person}{Uthaipon
  Tantipongpipat}, \bibinfo{person}{Jamie~H Morgenstern},
  \bibinfo{person}{Mohit Singh}, {and} \bibinfo{person}{Santosh Vempala}.}
  \bibinfo{year}{2018}\natexlab{}.
\newblock \showarticletitle{The price of fair PCA: One extra dimension}. In
  \bibinfo{booktitle}{\emph{Advances in Neural Information Processing
  Systems}}. \bibinfo{pages}{10976--10987}.
\newblock


\bibitem[\protect\citeauthoryear{Shore, Randel, Chung, Dean, Holcombe~Ehrhart,
  and Singh}{Shore et~al\mbox{.}}{2011}]%
        {inclusion1}
\bibfield{author}{\bibinfo{person}{Lynn~M Shore}, \bibinfo{person}{Amy~E
  Randel}, \bibinfo{person}{Beth~G Chung}, \bibinfo{person}{Michelle~A Dean},
  \bibinfo{person}{Karen Holcombe~Ehrhart}, {and} \bibinfo{person}{Gangaram
  Singh}.} \bibinfo{year}{2011}\natexlab{}.
\newblock \showarticletitle{Inclusion and diversity in work groups: A review
  and model for future research}.
\newblock \bibinfo{journal}{\emph{Journal of management}} \bibinfo{volume}{37},
  \bibinfo{number}{4} (\bibinfo{year}{2011}), \bibinfo{pages}{1262--1289}.
\newblock


\bibitem[\protect\citeauthoryear{Singh and Joachims}{Singh and
  Joachims}{2017}]%
        {fairness3}
\bibfield{author}{\bibinfo{person}{Ashudeep Singh} {and}
  \bibinfo{person}{Thorsten Joachims}.} \bibinfo{year}{2017}\natexlab{}.
\newblock \showarticletitle{Equality of opportunity in rankings}. In
  \bibinfo{booktitle}{\emph{Workshop on Prioritizing Online Content (WPOC) at
  NIPS}}.
\newblock


\bibitem[\protect\citeauthoryear{Snell}{Snell}{2017}]%
        {tokenism}
\bibfield{author}{\bibinfo{person}{Tonie Snell}.}
  \bibinfo{year}{2017}\natexlab{}.
\newblock \showarticletitle{Tokenism: The Result of Diversity Without
  Inclusion}.
\newblock \bibinfo{journal}{\emph{Medium}} (\bibinfo{year}{2017}).
\newblock
\urldef\tempurl%
\url{https://medium.com/@TonieSnell/tokenism-the-result-of-diversity-without-inclusion-460061db1eb6}
\showURL{%
\tempurl}


\bibitem[\protect\citeauthoryear{Tuomisto}{Tuomisto}{2011}]%
        {ecology_commentary}
\bibfield{author}{\bibinfo{person}{Hanna Tuomisto}.}
  \bibinfo{year}{2011}\natexlab{}.
\newblock \showarticletitle{Commentary: do we have a consistent terminology for
  species diversity? Yes, if we choose to use it}.
\newblock \bibinfo{journal}{\emph{Oecologia}} \bibinfo{volume}{167},
  \bibinfo{number}{4} (\bibinfo{year}{2011}), \bibinfo{pages}{903--911}.
\newblock


\bibitem[\protect\citeauthoryear{Whittaker}{Whittaker}{1960}]%
        {ecology2}
\bibfield{author}{\bibinfo{person}{Robert~Harding Whittaker}.}
  \bibinfo{year}{1960}\natexlab{}.
\newblock \showarticletitle{Vegetation of the Siskiyou mountains, Oregon and
  California}.
\newblock \bibinfo{journal}{\emph{Ecological monographs}} \bibinfo{volume}{30},
  \bibinfo{number}{3} (\bibinfo{year}{1960}), \bibinfo{pages}{279--338}.
\newblock


\bibitem[\protect\citeauthoryear{Yang and Stoyanovich}{Yang and
  Stoyanovich}{2017}]%
        {fairness2}
\bibfield{author}{\bibinfo{person}{Ke Yang} {and} \bibinfo{person}{Julia
  Stoyanovich}.} \bibinfo{year}{2017}\natexlab{}.
\newblock \showarticletitle{Measuring fairness in ranked outputs}. In
  \bibinfo{booktitle}{\emph{Proceedings of the 29th International Conference on
  Scientific and Statistical Database Management}}. ACM, \bibinfo{pages}{22}.
\newblock


\bibitem[\protect\citeauthoryear{Zhou, Kuscsik, Liu, Medo, Wakeling, and
  Zhang}{Zhou et~al\mbox{.}}{2010}]%
        {11}
\bibfield{author}{\bibinfo{person}{Tao Zhou}, \bibinfo{person}{Zolt{\'a}n
  Kuscsik}, \bibinfo{person}{Jian-Guo Liu}, \bibinfo{person}{Mat{\'u}{\v{s}}
  Medo}, \bibinfo{person}{Joseph~Rushton Wakeling}, {and}
  \bibinfo{person}{Yi-Cheng Zhang}.} \bibinfo{year}{2010}\natexlab{}.
\newblock \showarticletitle{Solving the apparent diversity-accuracy dilemma of
  recommender systems}.
\newblock \bibinfo{journal}{\emph{Proceedings of the National Academy of
  Sciences}} \bibinfo{volume}{107}, \bibinfo{number}{10}
  (\bibinfo{year}{2010}), \bibinfo{pages}{4511--4515}.
\newblock


\end{thebibliography}

\end{document}